\documentclass[5p,times]{elsarticle}

\usepackage{amssymb}
\usepackage{amsmath}

\journal{Nuclear Physics B}

\usepackage{tabularx}

\begin{document}

\begin{frontmatter}



\title{An Amphibious Untethered Inchworm Soft Robot for Fast Crawling Locomotion} 



\author[carleton]{Mohammadjavad Javadi}
\author[carleton]{Charlie Wadds}
\author[elixir]{Robin Chhabra\corref{cor1}}

\address[carleton]{Department of Mechanical and Aerospace Engineering, Carleton University, Ottawa, ON, Canada}
\address[elixir]{Embodied Learning and Intelligence for eXploration and Innovative soft Robotics (ELIXIR) Lab\\ Department of Mechanical, Industrial, and Mechatronics Engineering, Toronto Metropolitan University, Toronto, ON, Canada}

\cortext[cor1]{Corresponding author: robin.chhabra@torontomu.ca\\
No potential conflict of interest was reported by the author(s).\\
This work is partially supported by a grant from the Natural Sciences and Engineering Research Council of Canada (DGECR-2019-00085).}

\begin{abstract}
Untethered soft robots are essential for advancing the real-world deployment of soft robotic systems in diverse and multitasking environments. Inspired by soft-bodied inchworm, we present a fully untethered soft robot with a curved, flexible structure actuated by magnetic forces. The robot has a total mass of $102.63\,\mathrm{g}$ and demonstrates multimodal locomotion, achieving a maximum walking speed of $3.74\,\mathrm{cm/s}$ and a swimming speed of $0.82\,\mathrm{cm/s}$. A compact and lightweight onboard control circuit enables wireless command transmission, while an integrated camera provides environmental perception. Through structural optimization and system-level integration, the robot successfully performs walking, steering, swimming, and payload transport without reliance on external infrastructure. The robot’s dynamic performance and locomotion capabilities are systematically validated through experimental characterization.
\end{abstract}




\end{frontmatter}




\section{Introduction}
\begin{figure}[h]
  \begin{center}
  \includegraphics[width=3in]{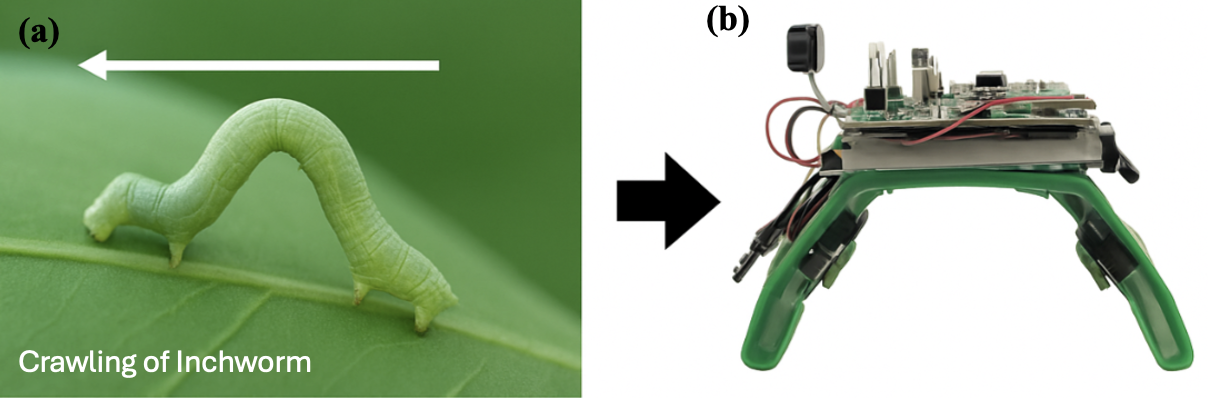}\\
  \caption{Bioinspiration of the proposed robot. (a) An inchworm in crawling motion \cite{zhang2024}. (b) The developed untethered soft robot, designed with a curved flexible body to
mimic inchworm's crawling locomotion.}
  \label{figg0}
  \end{center}
\end{figure}

In nature, locomotion is enabled by deformable bodies, inspiring soft robots that mimic biological motion and functionality. For example, animals such as the inchworm (Figure~\ref{figg0}a) navigate complex environments by bending their flexible bodies. Inspired by this mechanism, soft crawling robots have been developed for their ability to interact delicately with the environment~\cite{pan2025}.
Recently, roboticists have developed various types of soft robots made from hyper-flexible materials, enabling them to continuously change their shape or properties to adapt to different environments. This adaptability allows them to navigate and manipulate objects safely, even in confined spaces \cite{c1,c2}.
Soft robots have been employed in a variety of applications, including surgery \cite{c3,c4}, wearable rehabilitation devices \cite{c5}, drug delivery \cite{c26,c27,c28}, gripper design \cite{c6,c29}, and space exploration \cite{c16,c17,c18}. However, in most of these applications, soft robots often require a tethered connection to support pneumatic or electrical hardware and control systems.

The literature presents several innovative designs for untethered crawling soft robots. For example, Tolley \textit{et al.}~\cite{c19} developed an untethered soft robot that combines pneumatic and explosive actuators to perform directional jumping maneuvers, enabling it to overcome high obstacles through dynamic leaps. 
In the context of large-scale soft untethered robots, Usevitch \textit{et al.}~\cite{c21} reported an inflatable robotic truss composed of thin-walled, air-filled tubes. The structure is capable of shape morphing by continuously repositioning its joints while maintaining a constant total edge length. The robot demonstrated locomotion using a punctuated rolling gait, achieving a speed of 2.14 body lengths per minute. Cao \textit{et al.}~\cite{c22} developed an untethered soft robot powered by soft electrostatic actuators, capable of walking on flat surfaces at a speed of 0.02 body lengths per second.
Inspired by the locomotion of geckos, tree frogs, and inchworms, Sun \textit{et al.}~\cite{c23} designed an untethered soft robot capable of climbing walls and ceilings with various surface textures, including glass, polyimide, and aluminum.

In the case of walking soft robots, developing an untethered system using available mechanisms remains a challenging and complex problem in robotics. In this regard, Tolley \textit{et al.} developed an untethered, pneumatically powered soft robot with four legs and embedded chambers, enabling it to walk across different environments \cite{c8}. While the robot demonstrated the ability to navigate obstacles and tolerate damage, its pneumatic actuation mechanism resulted in slow movement $0.5 cm/s$.
Kaarthik \textit{et al.}~\cite{c20} designed a four-legged soft robot using 3D-printed cylindrical handed shearing auxetics (HSAs), fabricated from a single-cure polyurethane resin. The design is scalable for use in walking robots. Their robot demonstrated sustained locomotion at a speed of two body lengths per minute for up to 65 minutes and was capable of carrying a payload of 1.5~kg. Actuation forces were applied to the soft components through rotating motors.

\begin{figure}[h]
  \begin{center}
  \includegraphics[width=2.5in]{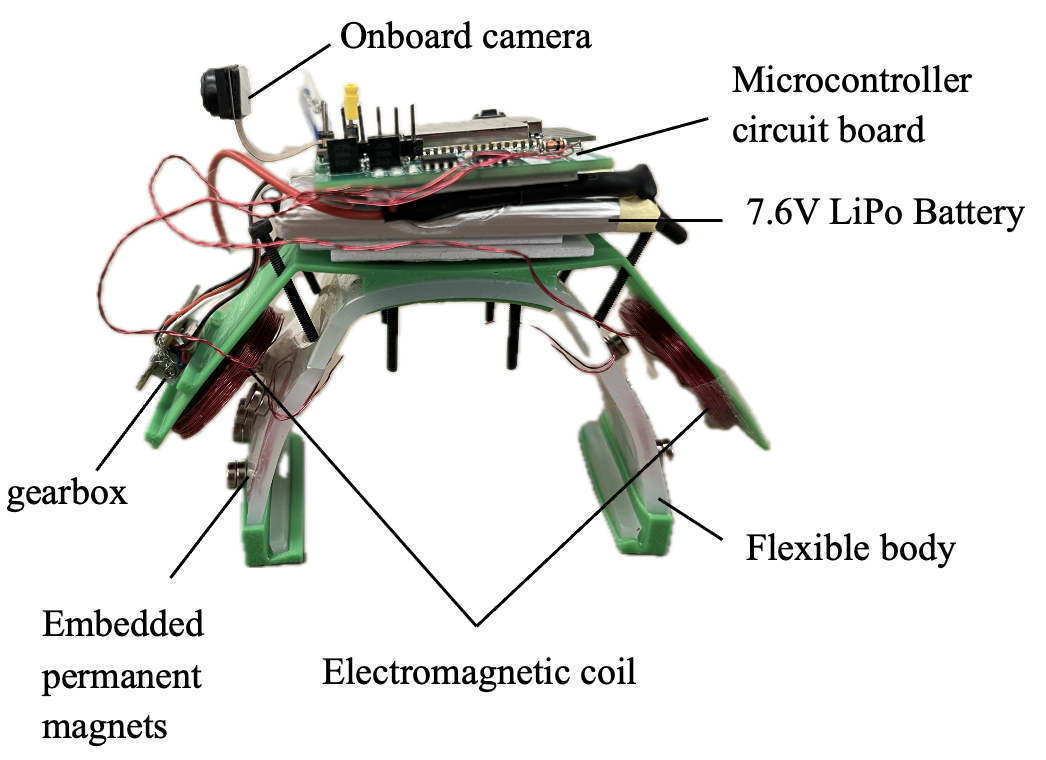}\\
  \caption{Components of the shape-morphing magnetic soft robot.}
  \label{figg1}
  \end{center}
\end{figure}

Huang \textit{et al.} proposed a novel 25g soft electrically actuated quadruped (SEAQ) robot capable of crawling at a maximum speed of 3.2 cm/s ($0.56$BL/s or body length per second) and adapting to different environments. This robot was faster than previous pneumatically actuated soft robots. However, one of the main limitations of SEAQ robots is their inability to produce continuous locomotion due to the actuation mechanism, which requires a cooldown period. This limitation further restricts the robot's operation in warm environments \cite{c9}. Mao \textit{et al.} proposed a novel small-scale, high-speed soft curved elastomeric robot driven by the Lorentz force acting on an embedded printed liquid metal. While the design indicates that the robot is untethered, it requires proximity to a strong external magnetic field \cite{c10}. 
Among the available mechanisms and active actuators for developing untethered soft robots, none has been actuated by a magnetic field.
In this study, however, we propose for the first time a fully untethered electromagnetic soft robot (see Figure~\ref{figg0}b), which not only crawls on the ground using a bioinspired locomotion mechanism but also carries its own magnetic actuator, in contrast to other existing models~\cite{c10}.
The details of the proposed soft robot are shown in Figure~\ref{figg1}, where the robot weighs $102.63\,\text{g}$ and achieves a maximum walking speed of $3.74\,\text{cm/s}$ ($0.5\,\text{BL/s}$).
The main body of the robot is made of a hyper-flexible material shaped like a half-cylinder, with a bilayer elastomer embedded with strong permanent magnets. On top of this structure, a lightweight chassis is attached to house the control circuits, a high-amperage battery, and strong magnetic coils. To change the robot's direction, a small servo motor is incorporated to control the external magnetic field and generate a non-uniform distribution. 
The main novelty of the proposed model lies in the incorporation of magnetic coils into the robot, providing an efficient design that enables it to carry its own coils. This feature makes the robot independent of external factors and distinguishes it from other tethered magnetic robots.
As demonstrated in this paper, the robot can continuously locomote across different surfaces, carry cargo, and even swim in water. To enhance its speed and stability, specially designed ``shoes'' are attached to its front and back legs. The primary contribution of this work is the development of an untethered soft robot powered by electromagnetic actuation, capable of carrying its own magnetic actuators while achieving fast, uniform, and multifunctional motion.

This paper is organized as follows: in Section~\ref{DF}, the design and fabrication process of the soft curved shell, the fabrication molds, the magnetic coil, and the assembly process are described. In Section~\ref{EC}, the lightweight and compact control board design, along with the gait design, is presented. Section~\ref{SRI} presents the experimental tests of the robots, including walking on different surfaces and swimming in water, with details about the appropriate frequency for each scenario. Section~\ref{FI} discusses the applications and potential future improvements of the robots. Finally, the conclusion is provided in Section~\ref{Con}.

\section{Design and Fabrication}\label{DF}
\begin{figure}
  \begin{center}
  \includegraphics[width=3in]{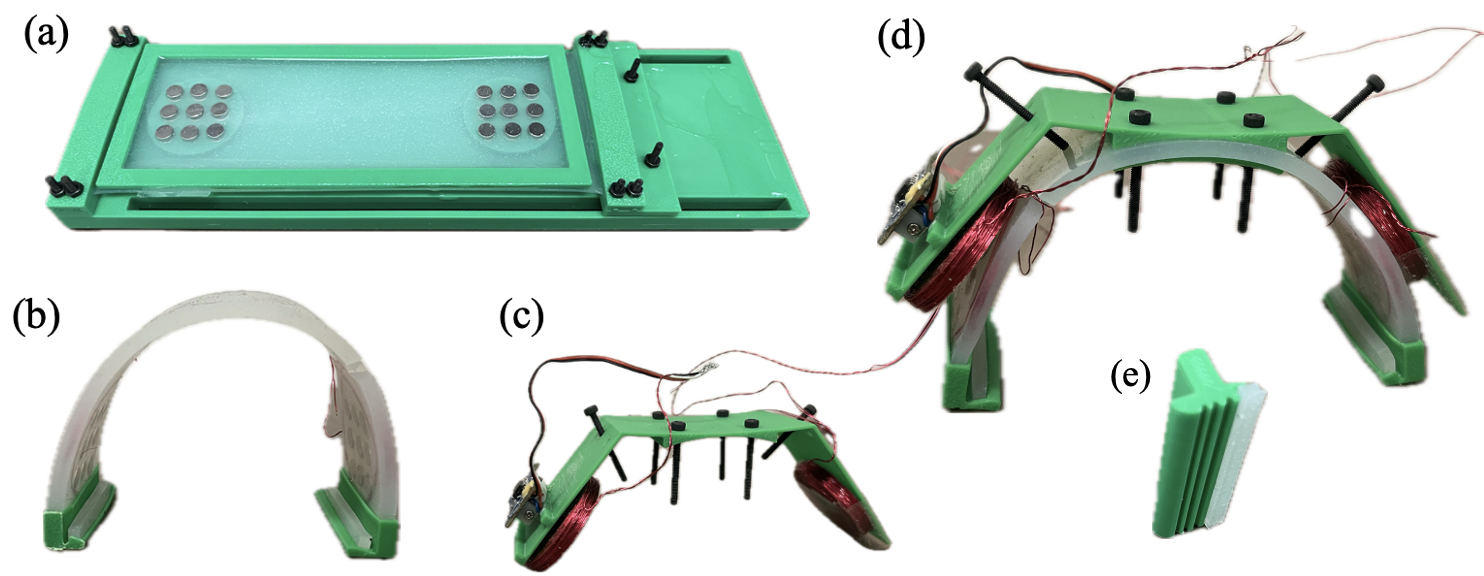}\\
  \caption{Fabrication of the soft magnetic robot. Fabrication schematic: (a) elastomer cured on pre-stretched layer with magnets on front and back legs. (b) Soft body with attached shoes. (c) Chassis with coil and a servo motor on one side. (d) Assembled robot. (e) Shoe design for improved stability.
}\label{fig1}
  \end{center}
\end{figure}

Referring to Figure \ref{fig1}, the magnetic soft robot consists of a soft bilayer curved elastomer with embedded permanent magnets, a rigid chassis with attached electronic circuits, a battery, and electromagnetic coils. The permanent magnets are affixed to the front and back legs of the robot, while the electromagnetic coils are mounted on the chassis above them to facilitate actuation. To maneuver the robot and change its direction, a small gearbox is installed on one side to adjust the position of the coil by moving it to the left or right. The gearbox employs a helical rack-and-pinion system, which efficiently converts the rotational motion of the pinion (a round gear) into precise linear motion along the rack (a straight gear). For improved interaction with the ground and enhanced stability during motion, the tips of the legs are equipped with appropriately designed shoes.  
A 3D-printed shoe, shown in Figure~\ref{fig1}e, with a bristled structure and a soft component that provides anisotropic friction characteristics for crawling.
A small-scale printed circuit board (PCB) containing a microcontroller, an Inertial Measurement Unit (IMU), a drive circuitry, and a camera along with an attached battery, is provided for controlling and navigating the robot. The battery weighs $26\,\mathrm{g}$, and the weight of the PCB is $10.64\, \mathrm{g}$. When a step-function current is applied to the embedded coils, the resulting external magnetic field generates a bending moment on the soft parts of the robot. This deformation, together with the difference in friction forces at the contact points of the robot’s front and back legs, alters its shape and shifts its center of mass, enabling it to walk across the ground. Due to the elastic nature of the soft body components, the legs return to their initial configuration without significant energy dissipation.
As shown in Figure \ref{fig1}b, the flexible bilayer part is made of Mold Star 31T, with a Young's modulus of $E = 324.054$ kPa. 
By actuating the permanent magnet with an on/off control current applied to the attached magnetic coil, we generate a maximum magnetic field in the range ($0mT - 22mT$), enabling the forward and backward movement of the soft robot's legs.

\subsection{Soft body fabrication}
The main body of the robot possesses the following characteristics:  
(i) it is sufficiently soft to adapt to varying environmental conditions and external impacts while also being robust enough to support the weight of the electronics and battery,  
(ii) it can rapidly generate adequate deformations to facilitate movement on the ground or swimming in water,  
(iii) it enables fast motion and exhibits tolerance to damage,  
(iv) it can fold and twist to change direction and maneuver the robot.  
This is achieved by bonding two layers of a two-part prepolymer, which consists of a mixture of Part A and Part B of Mold Star 31T in a $1:1$ mass ratio through a uniform mixing process. The elastomer is then partially cured at room temperature for $23$ minutes.

Following the computational analysis presented in \cite{c10}, the curved shell is designed by first preparing a flat plate with dimensions $16 \text{ cm} \times 5 \text{ cm}$ and a thickness of $1 \text{ mm}$ using Mold Star 31T. After curing, the layer is stretched by $5 \text{ mm}$ and fixed in the designated fixture, as shown in Figure \ref{fig1}a. A set of nine circular permanent magnets, each with a diameter of $5\,\text{mm}$, a thickness of $1.7\,\text{mm}$, and a magnetic field strength of $125\,\text{mT}$, is positioned on top of this layer. To temporarily fix the magnets before adding another layer, two circular holders with nine holes are designed and placed beneath the first layer, allowing another set of magnets to be aligned on top. After curing, the holders are removed. Once the permanent magnet is properly positioned, another mold with dimensions $13 \text{ cm} \times 4 \text{ cm}$ and a thickness of $2 \text{ mm}$ is placed on top of the layer, followed by pouring Mold Star 31T again. After curing, the layer is separated from the fixture, as illustrated in Figure \ref{fig1}b.  

The final shape of the robot's body, shown in Figure \ref{fig1}d, is a half-circular shell with a thickness of $3 \text{ mm}$, a length of $35 \text{ mm}$, and a diameter of $81.53 \text{ mm}$. This diameter represents the distance between the front and back legs of the robot in its relaxed configuration.
\subsection{Magnetic field characterization}
To generate the magnetic field, two circular magnetic coils were fabricated using 30 AWG copper wire with a diameter of 0.2 mm. By winding the wire around a cylindrical core, two functional magnetic coils are produced. The coil attached to the gearbox is capable of generating a magnetic field of 21.07~mT, with a maximum pulling/pushing force of 0.25~N on the leg. The other coil generates 19.4~mT with a maximum pulling/pushing force of 0.2~N.
The coil mounted on the gearbox weighs 11.91~g, while the coil on the opposite side weighs 13.5~g. The forces are measured using a digital force gauge (Push Pull Gauge Portable Force Meter) with an error margin of $\pm 1\%$.

\begin{figure}
  \begin{center}
  \includegraphics[width=3in]{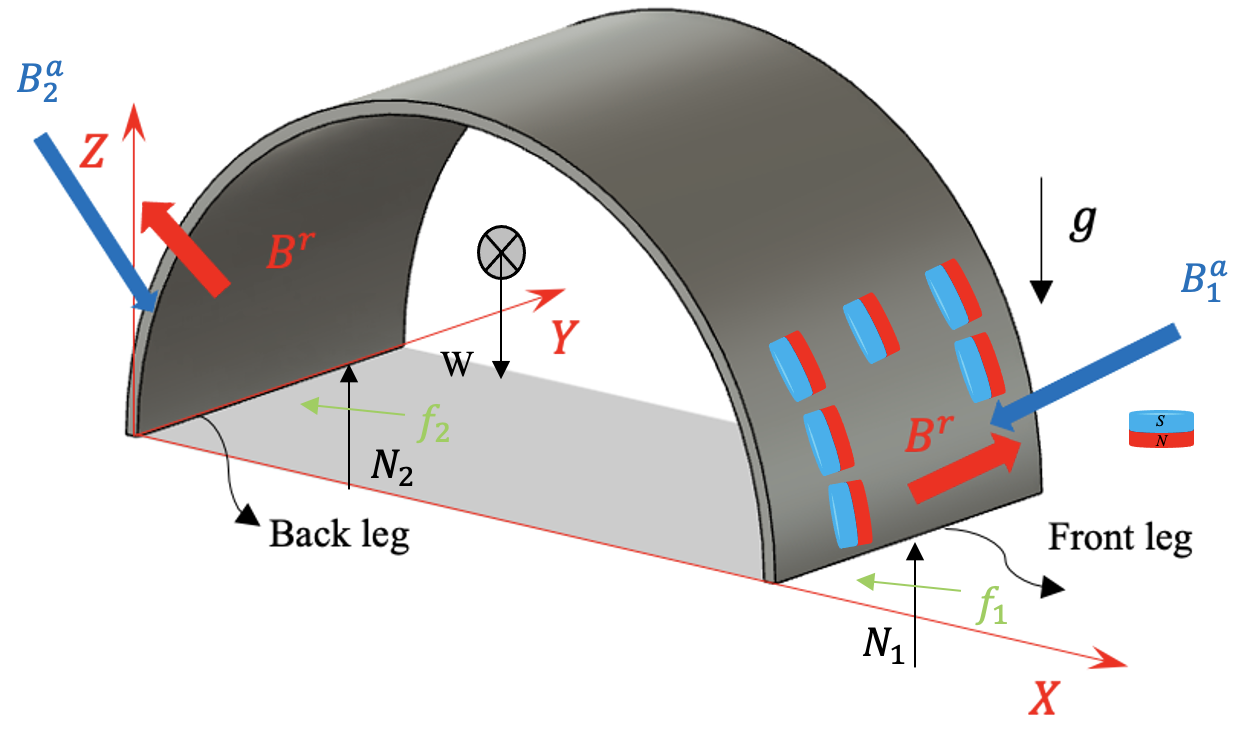}\\
  \caption{Schematic of the soft robot under applied magnetic fields.
.}\label{fig2}
  \end{center}
\end{figure}

If we denote the (remanent) magnetization field in the embedded permanent magnet by $\mathbf{B}^{r}$, then under the influence of an externally applied magnetic field due to the coils, $\mathbf{B}^{a}$, their interaction can be expressed in terms of a body force per unit volume as  
\begin{align}\label{E1}
\mathbf{f} = \dfrac{1}{\mu_{0}} \mathbf{B}^{r} \cdot \nabla \mathbf{B}^{a},
\end{align}  
and a body couple per unit volume as  
\begin{align}\label{E2}
\mathbf{m} = \dfrac{1}{\mu_{0}} \mathbf{B}^{r} \times \mathbf{B}^{a},
\end{align}  
where $\mu_{0}$ is the permeability of free space, and $\nabla$ is the spatial gradient operator~\cite{c12}.  
Here, both $\mathbf{B}^{r}$ and $\mathbf{B}^{a}$ are vector fields. The gradient of the applied field, $\nabla \mathbf{B}^{a}$, is a second-order tensor. The operator ``$\cdot$'' in Eq.~\eqref{E1} denotes the inner product between $\mathbf{B}^{r}$ and $\nabla \mathbf{B}^{a}$, while the operator ``$\times$'' in Eq.~\eqref{E2} represents the cross product (outer product) between $\mathbf{B}^{r}$ and $\mathbf{B}^{a}$. If the applied magnetic field is uniform, then its gradient vanishes, i.e., $\nabla \mathbf{B}^{a} = \mathbf{0}$. The gradient can be evaluated with respect to an inertial coordinate frame.  
The applied force and couple induce the actuation of the robot's legs during motion. A schematic representation of the applied loads on the robot is shown in Figure~\ref{fig2}.

\section{Electronic Circuits}\label{EC}

\subsection{Electrical design}

\begin{figure}
  \begin{center}
  \includegraphics[width=3in]{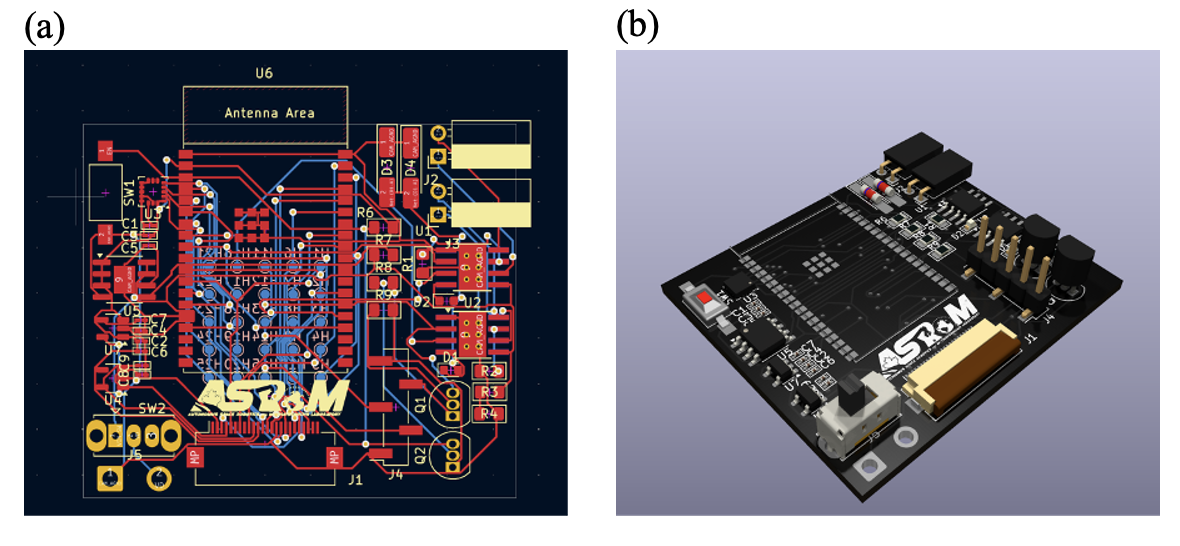}\\
  \caption{Schematic of the PCB: (a) PCB layout diagram showing the placement and routing of components; (b) 3D rendered view of the PCB illustrating the physical appearance of the assembled board.}
\label{figpcb}
  \end{center}
\end{figure}  


To enable untethered operation while maintaining a lightweight design, we developed a custom PCB that integrates a microcontroller, wireless communication hardware, power management circuitry for electromagnets, an inertial measurement unit (IMU), and an optional camera (see Figure \ref{figpcb}). The design prioritizes both functionality and ease of development and is light enough to significantly increase the performance of untethered operations.
The PCB features an ESP32 microcontroller, chosen for its dual-core architecture that permits one core to handle high-rate sensor readings while the other manages communication and control tasks \cite{c32}. The ESP32 supports programming via the Arduino framework for rapid prototyping and the more customizable ESP-IDF for performance-critical applications. Its integrated WiFi and Bluetooth modules enable remote control, wireless programming, and other advanced functionalities. In addition, onboard flash memory supports extensive data logging.
For driving the magnetic coils, the board employs two Texas Instruments DRV8872 H-Bridge motor drivers, selected for their lightweight design, high current capability, built in protection features, and ease of use \cite{c33}. The H-Bridges are controlled via PWM-capable pins on the ESP32, enabling precise modulation of the power applied to each coil and, consequently, a wide range of achievable gaits.
The design also incorporates a reset button and dedicated serial programming pins that are compatible with FTDI-based programmers and includes the required programming logic signals, streamlining the flashing process with automatic reset flags. Voltage regulation is achieved through a suite of regulators (3.3 V, 2.8 V, and 2.2 V), ensuring stable logic-level voltages for all integrated components.

\subsection{Soft robot gait}
\begin{figure}
  \begin{center}
  \includegraphics[width=2.5in]{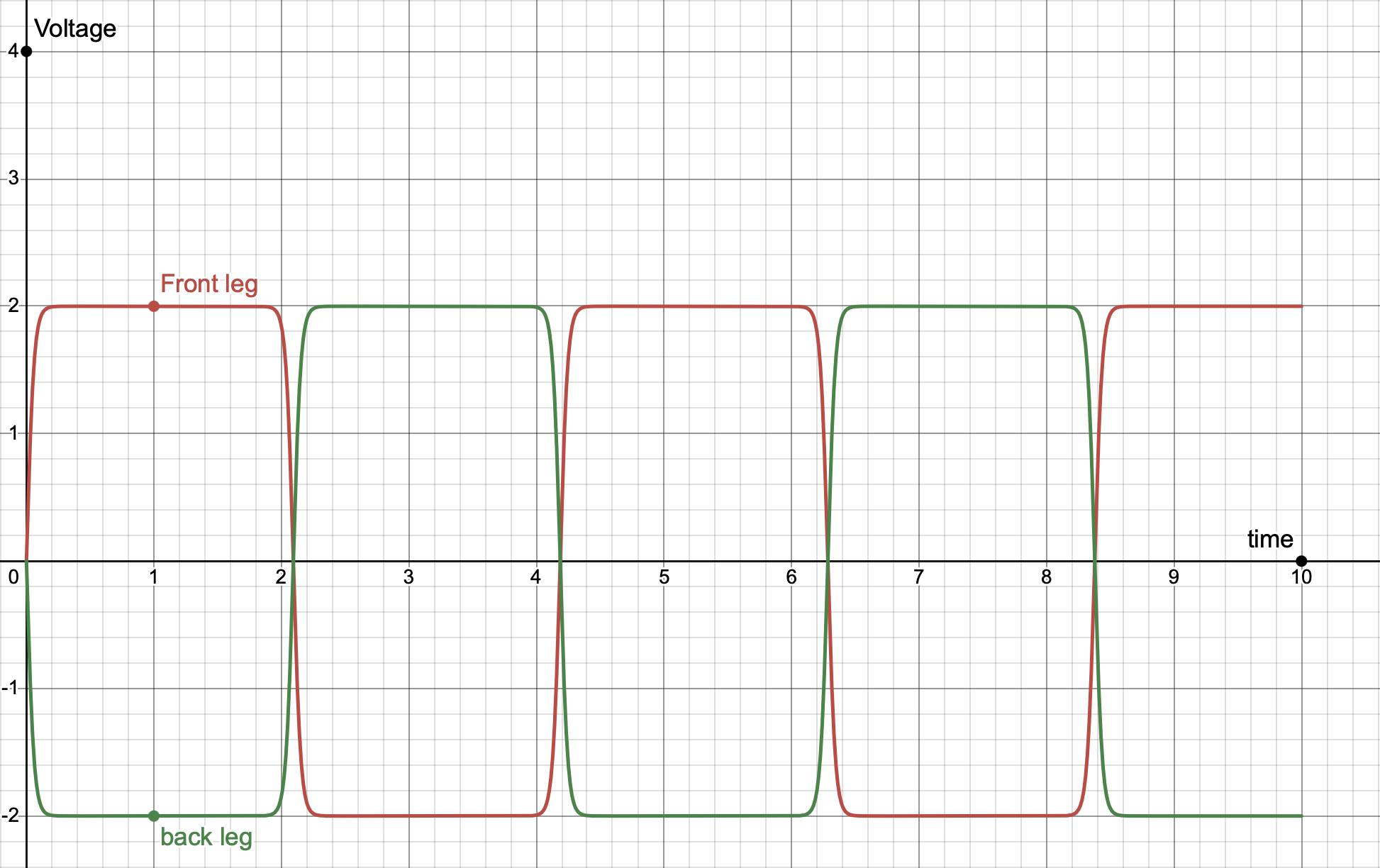}\\
  \caption{Illustration of the gait used. This represents an arbitrary frequency and amplitude and is used to show the motion of the legs only.
.}\label{figgait}
  \end{center}
\end{figure}  

The robot's gait is configured in software using hardware-based timers on the ESP32 microcontroller. The gait used during terrestrial walking tests is inspired by quadrupedal animal locomotion \cite{c34}, where the front and rear legs operate 180 degrees out of phase (see Figure \ref{figgait}). For aquatic testing, the gait is modified to a unidirectional, in-phase motion in which the actuators moved in one direction and then passively returned to the neutral position. This modification is implemented to minimize reverse hydrostatic forces exerted by the water on the leg surfaces. In the original walking gait, these opposing forces tend to cancel out net movement, resulting in the robot remaining stationary when submerged. The gait frequency is also configurable in software, allowing for adaptation to different environments. Future work could involve real-time gait optimization based on sensor inputs such as camera or IMU data, potentially enabling higher speeds and improved locomotion across a broader range of surfaces.

\section{Soft Robot Implementation}\label{SRI}
This section presents the experimental evaluation of the proposed untethered soft robot in different scenarios, including locomotion on flat and inclined surfaces, turning, cargo transport, and swimming. A large set of experiments was performed to systematically investigate the robot's performance, and the results are refined into clear trends and key conclusions. A summary of the main findings is provided in Table~\ref{tab:summary}.

\subsection{Locomotion on flat and inclined surfaces}
\begin{figure}[h]
  \centering
  \includegraphics[width=3in]{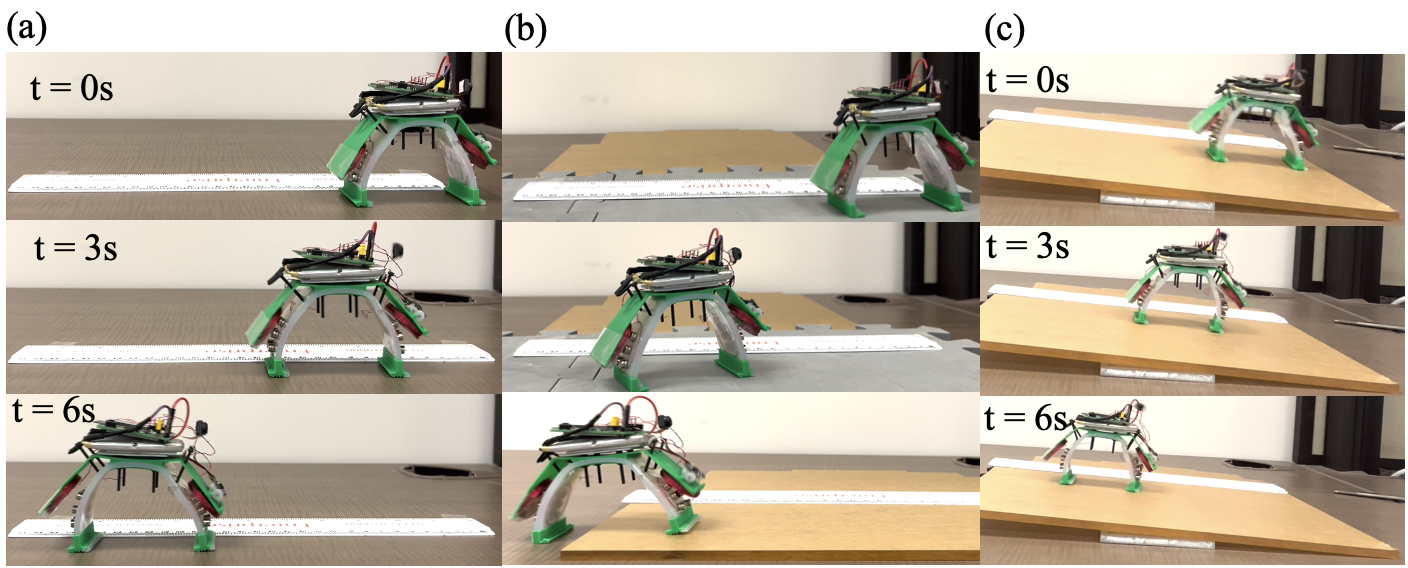}
  \caption{Walking behavior of the soft magnetic robot: (a) Schematic of locomotion on a flat plastic table at 4~Hz; (b) Locomotion on foam and paper surfaces; (c) Locomotion on a steep inclined surface.}
  \label{figwalk}
\end{figure}

\begin{figure}[h]
  \centering
  \includegraphics[width=3in]{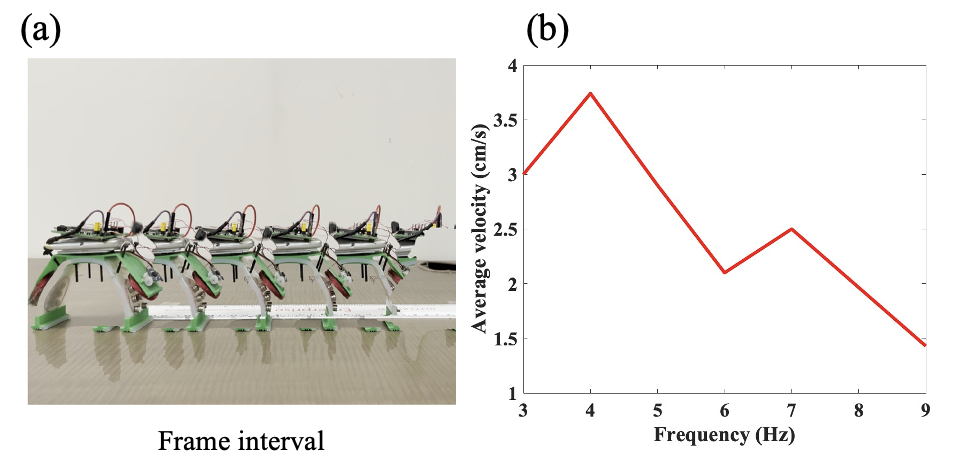}
  \caption{(a) Composite image of 4~Hz walking, frames extracted at 90 intervals. (b) Average velocity versus actuation frequency on a plastic table.}
  \label{figFV}
\end{figure}

\begin{figure}[h]
  \centering
  \includegraphics[width=2in]{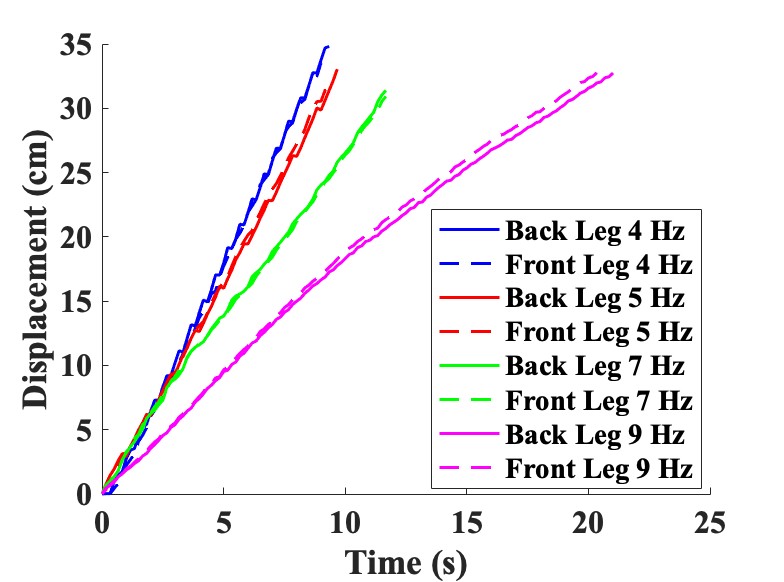}
  \caption{Displacement of the robot's front and back legs at various actuation frequencies during locomotion on a plastic table.}
  \label{figdis}
\end{figure}

\begin{figure}[h]
  \centering
  \includegraphics[width=2.5in]{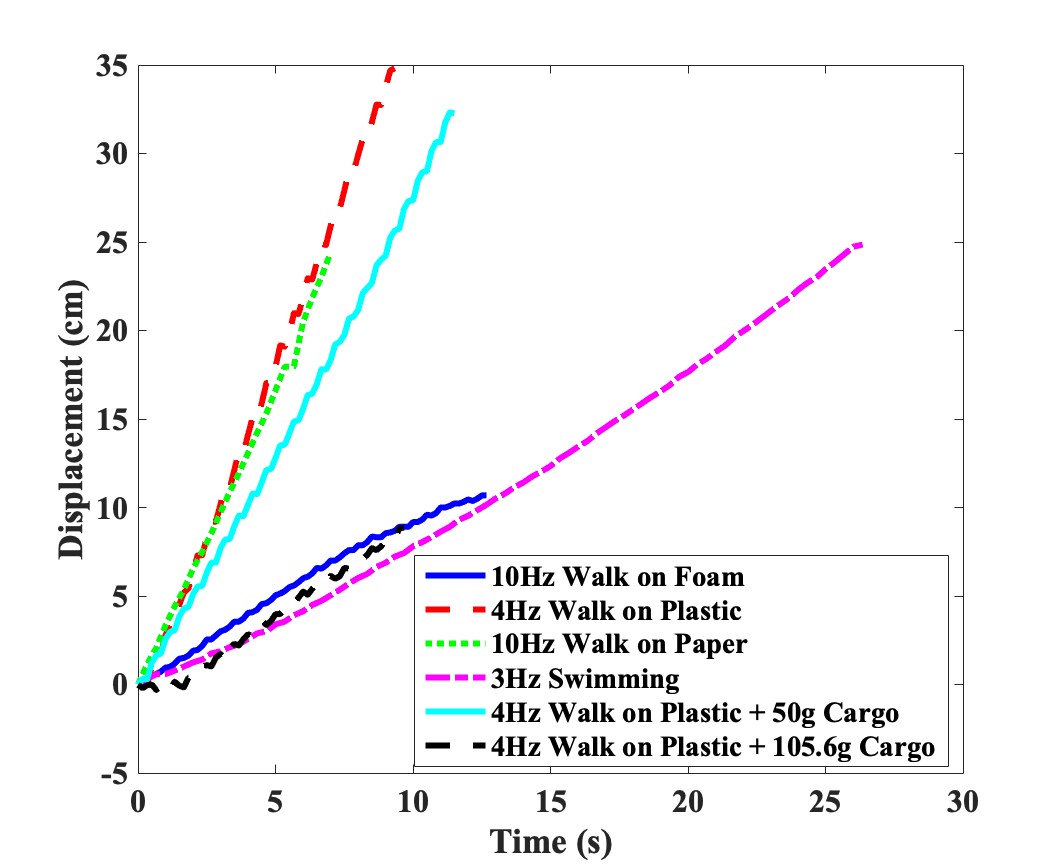}
  \caption{Center of mass displacement across different terrains and swimming conditions.}
  \label{figT}
\end{figure}

In the locomotion scenario, the robot consists of only two legs. To regulate the contact surface and ensure stable motion, specially designed shoes were attached to the legs (see Figure~\ref{fig1}e). The locomotion gait was generated by actuating the two legs with different amplitudes, creating an asymmetric load distribution that deforms the robot’s curved arc and shifts its center of mass. This results in a controlled variation in friction forces, enabling forward movement. 
Figure~\ref{figwalk} demonstrates the walking performance of the robot on surfaces with different friction coefficients. The same gait was applied across all surfaces, with the actuation frequency varied to adapt to surface conditions. The robot achieved its maximum speed of $3.74$\,cm/s on a plastic table at $4$\,Hz (see Supplementary Video~S1). As friction increased, higher actuation frequencies were required for locomotion; for instance, walking on foam required $10$\,Hz (see Video~S2). The displacement of the center of mass for different surfaces is compared in Figure~\ref{figT}. 
The robot can also ascend and descend inclined surfaces with slopes up to $7^{\circ}$ (Supplementary Videos~S5--S6). Figure~\ref{figFV}b presents the velocity–frequency relationship, showing that speed increases with frequency up to $4$\,Hz, then decreases due to reduced stroke effectiveness. Displacement data of the robot’s legs at different frequencies is shown in Figure~\ref{figdis}, confirming smoother and more uniform motion at higher frequencies. Walking on common indoor surfaces, such as office tiles, was also achieved at $4$\,Hz (see Video~S14).

\subsection{Turning ability}
\begin{figure}[h]
  \centering
  \includegraphics[width=3in]{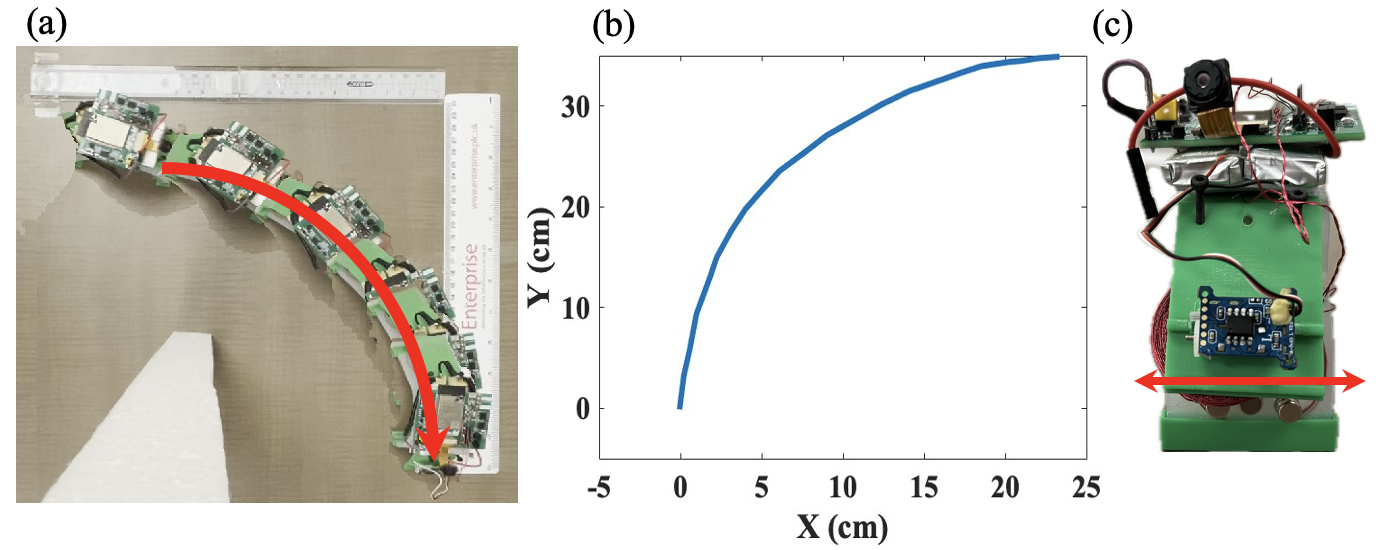}
  \caption{Turning behavior of the robot: (a) Composite image of right turning at 4~Hz; (b) Center of mass trajectory during right turn; (c) Schematic of lateral coil shift for turning.}
  \label{figTurn}
\end{figure}

Because the robot has only two legs, turning is inherently difficult. To address this limitation, a linear gearbox was integrated to shift the electromagnetic coil laterally (Figure~\ref{figTurn}c). Moving the coil left causes the robot to turn right, and vice versa. This occurs due to asymmetric bending moments, which induce twisting in addition to bending of the robot’s leg.
At 4~Hz, the robot achieved an average angular velocity of $0.087$\,rad/s, corresponding to a turning radius of approximately $28$\,cm (Figure~\ref{figTurn}a,b, Video~S7). 

\subsection{Cargo transport}
\begin{figure}[h]
  \centering
  \includegraphics[width=3in]{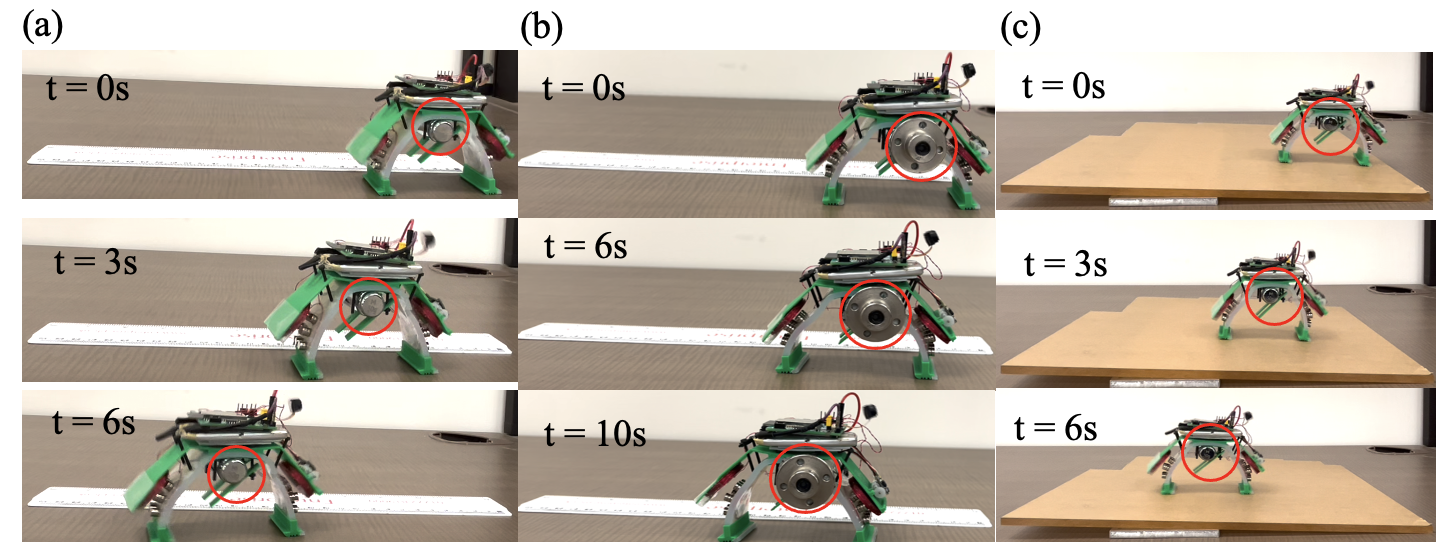}
  \caption{Cargo transport capability: (a) Locomotion with a 50~g load at 4~Hz; (b) Locomotion with a 105.6~g load; (c) Walking with cargo on inclined surface.}
  \label{figcargo}
\end{figure}

A major challenge in untethered soft robots is balancing self-weight with actuation strength. Our design supports its own weight and can carry significant cargo loads. Figure~\ref{figcargo} shows the robot transporting $50$\,g and $105.6$\,g payloads with average velocities of $2.5$\,cm/s and $0.8$\,cm/s, respectively (see also Figure~\ref{figT}). The robot maintained stable locomotion even on inclined surfaces while carrying cargo (Videos~S8--S10). 

\subsection{Swimming performance}
\begin{figure}[h]
  \centering
  \includegraphics[width=3in]{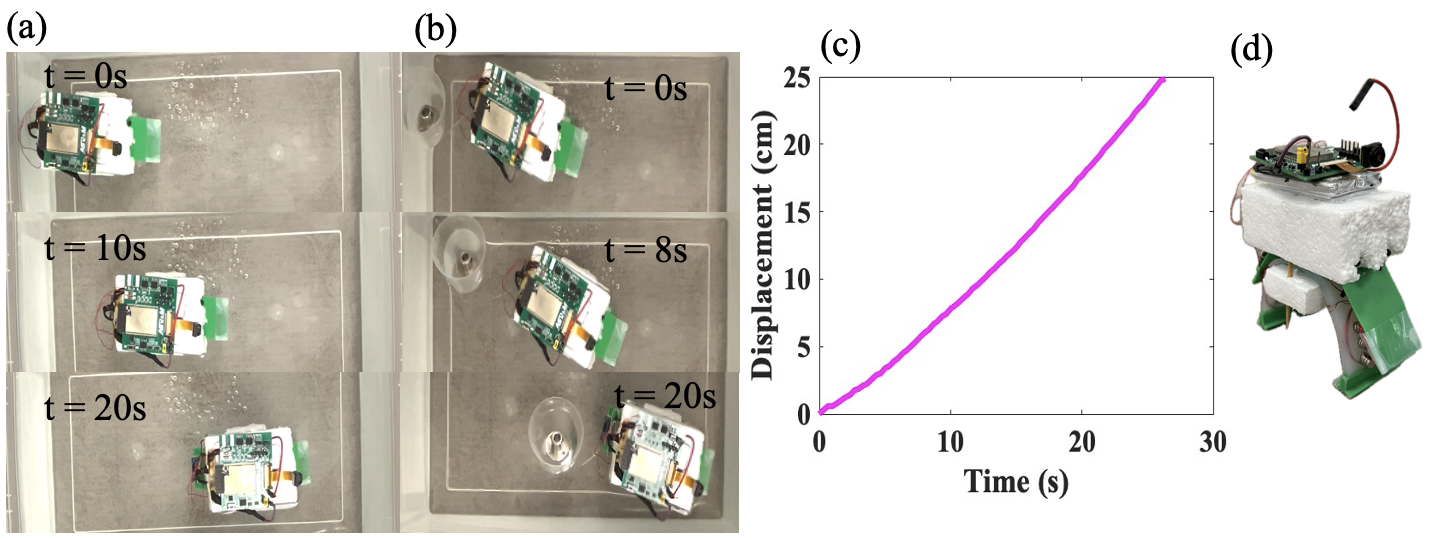}
  \caption{Swimming performance: (a) Swimming in water at 3\,Hz; (b) Swimming while pulling 8.8\,g cargo; (c) Center of mass displacement during swimming; (d) General schematic of the swimming robot.}
  \label{figswim}
\end{figure}

In addition to terrestrial locomotion, the robot can swim along the water surface (Videos~S11--S13). A lightweight foam block ensures buoyancy and houses the electronics above the waterline (Figure~\ref{figswim}d). The robot achieved a maximum swimming speed of $0.82$\,cm/s at $3$\,Hz. Figure~\ref{figswim}c shows its displacement over time. It is also capable of towing cargo between $8.8$\,g and $27.8$\,g (Figure~\ref{figswim}b, Video~S12). 

\subsection{Summary of experimental findings}
To emphasize the significance of the experimental results, Table~\ref{tab:summary} synthesizes performance across all scenarios, summarizing maximum speed, turning capability, and load-carrying performance. The experiments demonstrate that the untethered soft magnetic robot achieves versatile, robust locomotion in diverse environments. Key trends include: 
(1) Locomotion performance increases with actuation frequency up to an optimal $\sim$4\,Hz, beyond which stroke effectiveness and velocity decline;  
(2) The robot maintains stability while carrying loads exceeding its own body weight, highlighting actuation efficiency relative to structural mass;  
(3) Simple design modifications (coil-shift for turning, foam block for buoyancy) enable directional control and multimodal locomotion, including swimming and cargo transport.  
These findings confirm the adaptability and practicality of the design, establishing its potential as a lightweight, untethered platform for terrestrial and aquatic mobility.

\begin{table*}[h]
\centering
\caption{Summary of experimental performance of the soft magnetic robot.}
\label{tab:summary}
\resizebox{\textwidth}{!}{%
\begin{tabular}{|l|c|c|c|}
\hline
\textbf{Scenario} & \textbf{Condition} & \textbf{Maximum Performance} & \textbf{Key Observation} \\
\hline
Locomotion (flat surface) & Plastic table, 4 Hz & 3.74 cm/s & Peak velocity; smoother motion at higher frequency \\
\hline
Locomotion (foam) & Foam, 10 Hz & 1.9 cm/s (approx.) & Requires higher frequency due to friction \\
\hline
Inclined walking & Slope up to $7^{\circ}$ & 2.1 cm/s (approx.) & Stable ascent/descent achieved \\
\hline
Turning & Plastic table, 4 Hz & 0.087 rad/s ($\sim$28 cm radius) & Coil-shift enables directional control \\
\hline
Cargo transport & Flat surface, 4 Hz & 50 g: 2.5 cm/s; 105.6 g: 0.8 cm/s & Stable locomotion with heavy load \\
\hline
Swimming & Water surface, 3 Hz & 0.82 cm/s & Stable buoyant locomotion \\
\hline
Swimming with cargo & Water surface, 3 Hz & Up to 27.8 g cargo & Demonstrates towing capability \\
\hline
\end{tabular}
}
\end{table*}


\section{Future Improvements}\label{FI}

\subsection{Application}
The electromagnetic actuator forms the foundation of the robot’s multifunctionality. The robot can walk, explore its environment with an attached camera, swim due to the actuator’s water resistance, and carry small payloads. Its simple design allows easy assembly without precise adjustments, while a compact lightweight control board enables untethered operation. Experimental results highlight a novel use of electromagnetic actuators in soft untethered robotics, a direction not yet explored in the literature.

\subsection{Design and model}
Investigating electromagnetic mechanisms for multifunctional soft robots, while achieving higher walking speeds than existing fully untethered models, requires broader evaluation within the robotics community. A key limitation of the proposed robot is its restricted ability to jump or lift its legs. More flexible leg designs and optimized gait patterns are needed to enhance obstacle negotiation and stability. Incorporating compliant legs and reshaping feet could improve locomotion over uneven terrain, while repositioning embedded magnets or coils may yield more effective actuation.  
The current rigid chassis, though essential for housing coils and maintaining structure, reduces damage tolerance and compactness. Replacing rigid components with smart materials featuring tunable stiffness or shape memory alloys could improve resilience and self-recovery.  
The robot achieves approximately $126\,\text{cm}$ of locomotion in $90\,\text{s}$, after which the coils require a $2$--$3\,\text{min}$ cooldown, significantly outperforming the model in~\cite{c9}. Further improvements in circuit board and coil design---particularly resistance and voltage handling---are needed, with additional circuit enhancements discussed in the next section.

\subsection{Electrical circuits}
The current control board represents a compromise between lightweight performance and rapid prototyping. Exposed pins, through-hole components, and the relatively large ESP32 module contribute additional weight. Future designs could adopt a more compact architecture using an STM32 or equivalent microcontroller with SMD components on both sides of the PCB, reducing size and mass. Additionally, replacing the rigid PCB with a flexible PCB would allow embedding directly within the robot's silicone body, improving heat dissipation, enabling arbitrary sensor placement, and enhancing waterproofing—critical for more robust and versatile soft robots.

\section{Conclusion}\label{Con}

This study presents a novel, fully untethered electromagnetic soft robot demonstrating fast, multifunctional performance across multiple tasks. The two-leg design simplifies gait generation and achieves a maximum walking speed of $3.74$~cm/s, substantially outperforming comparable soft robotic systems reported in the literature. Locomotion is enabled by magnetic forces and torques acting on an embedded permanent magnet, allowing rapid body deformation and water-resistant operation. The robot can swim and steer in aquatic environments solely by modulating the external magnetic field. A custom miniaturized control board supports onboard wireless communication via Wi-Fi while maintaining a lightweight, compact form factor. The system also accommodates an integrated camera for environmental monitoring and successfully carries payloads exceeding its own body weight. These experimental results demonstrate that electromagnetic actuation provides a viable alternative to conventional pneumatic or electroactive methods for untethered soft robotic platforms. Future enhancements may include shape-morphing chassis structures or additional limbs to enable jumping and advanced obstacle negotiation.







\bibliographystyle{elsarticle-num}
\biboptions{square,sort,comma}
\bibliography{thebibliography}       

\end{document}